\documentclass[12pt]{article}
\usepackage{amsmath}
\usepackage{graphicx,psfrag,epsf}
\usepackage{enumerate}
\usepackage[sort&compress,comma,square,numbers]{natbib}
\usepackage{url}
\usepackage{thmtools, thm-restate}
\usepackage{enumitem}
\usepackage{booktabs}
\usepackage{amssymb, amsfonts,mathtools, amsthm}
\usepackage[colorlinks = true,
linkcolor = blue,
urlcolor  = blue,
citecolor = green,
anchorcolor = blue]{hyperref}
\usepackage[font=small,labelfont=bf]{caption}
\usepackage{authblk}
\usepackage{setspace}
\usepackage{lscape}
\usepackage{color}
\usepackage{dcolumn}
\usepackage{indentfirst, verbatim, float}
\usepackage{subcaption}
\usepackage{sidecap}
\usepackage{titlesec}

\newtheorem{definition}{Definition}

\providecommand{\sct}[1]{{\normalfont\textsc{#1}}}

\newcommand{\Dcor}{\sct{Dcor}}
\newcommand{\Dcov}{\sct{Dcov}}

\newcommand{\Hsic}{\sct{Hsic}}
\newcommand{\Hsicr}{\sct{HsiCor}}

\usepackage{helvet}

\begin{document}

\def\spacingset#1{\renewcommand{\baselinestretch}%
{#1}\small\normalsize} \spacingset{1}

\title{\bf The Exact Equivalence of Distance and Kernel Methods in Hypothesis Testing}
\author[1]{Cencheng Shen\thanks{shenc@udel.edu}}
\author[2,3]{Joshua T. Vogelstein\thanks{jovo@jhu.edu}}
\affil[1]{Department of Applied Economics and Statistics, University of Delaware}
\affil[2]{Institute for Computational Medicine, Johns Hopkins University}
\affil[3]{Department of Biomedical Engineering and Institute of Computational Medicine, Johns Hopkins University}
  \maketitle

\bigskip
\begin{abstract}
Distance correlation and Hilbert-Schmidt independence criterion are widely used for independence testing, two-sample testing, and many inference tasks in statistics and machine learning. These two methods are tightly related, yet are treated as two different entities in the majority of existing literature. In this paper, we propose a simple and elegant bijection between metric and kernel. The bijective transformation better preserves the similarity structure, allows distance correlation and Hilbert-Schmidt independence criterion to be always the same for hypothesis testing, streamlines the code base for implementation, and enables a rich literature of distance-based and kernel-based methodologies to directly communicate with each other.
\end{abstract}

\noindent%
\vfill
\spacingset{1.45} 
 
\section{Introduction}
\label{sec:intro}
Distance correlation is a distance-based method initially proposed for testing independence \citep{SzekelyRizzoBakirov2007,SzekelyRizzo2009}. It can be used in two-sample test \citep{RizzoSzekely2016,exact1}, conditional independence \citep{SzekelyRizzo2014,Wang2015}, feature screening \citep{LiZhongZhu2012, Zhong2015,mgc4}, clustering \citep{Szekely2005,Rizzo2010}, time-series testing \citep{Zhou2012,Pitsillou2018,mgc5}, graph dependence~\citep{mgc3,mgc6}. The Hilbert-Schmidt independence criterion is a kernel-based method for testing independence and equally popular in related inference tasks \citep{GrettonEtAl2005, Gretton2007,Song2007, GrettonGyorfi2010, bal2013,Chang2013, Zhang2018}. These two foundational methods are universally consistent for testing independence, which motivated many other consistent methods with improved finite-sample power \citep{HellerGorfine2013,heller2016consistent,mgc1,mgc2,Zhu2017,Zhu2018,Kim2018, mgc7}.

The distance-based and kernel-based methods share similar formulations and many common properties. The independence hypothesis is formulated as follows: given paired sample data $\{(x_{i},y_{i}) \in \mathbb{R}^{p+q}, i=1,\ldots,n \}$ where $p$ denotes the dimension of $x_i$ and $q$ denotes the dimension of $y_i$, we aim to test
\begin{align*}
&H_{0} : F_{XY} = F_{X}F_{Y}, \\
&H_{A} : F_{XY} \neq F_{X}F_{Y}
\end{align*}
by assuming $\{(x_{i},y_{i})\stackrel{\emph{iid}}{\sim} F_{XY}\}$. The sample distance correlation is computed via the two pairwise Euclidean distance matrices, is proved asymptotically $0$ if and only if $X$ and $Y$ are independent, and thus is universally consistent for testing independence. On the other hand, the Hilbert-Schmidt independence criterion is computed via the Gaussian kernel, is also proved asymptotically $0$ if and only if $X$ and $Y$ are independent, and thus is universally consistent for testing independence. 

Distance correlation is applicable to any metric choice, while the Hilbert-Schmidt independence criterion may use any kernel choice, and the consistency property still holds under proper conditions \cite{Lyons2013, GrettonEtAl2005}. It was shown in \cite{SejdinovicEtAl2013} that there exists a fixed-point transformation between metrics and kernels such that distance correlation is equivalent to the Hilbert-Schmidt independence criterion for the population statistic. However, this transformation requires an arbitrary fixed point to start with. Moreover, it is still an open question whether the equivalence holds between the sample statistics and between testing p-values, in particular the unbiased statistics \cite{Song2007,SzekelyRizzo2014}.

In this paper, we propose a new bijective transformation between metrics and kernels. The bijection is shown as a special case of the fixed-point transformation proposed in \cite{SejdinovicEtAl2013} but more advantageous for sample data. The bijection is intuitive to use and simpler to implement, does not rely on a pre-determined fixed point, enables testing equivalence between distance correlation and Hilbert-Schmidt independence criterion using either the biased or the unbiased statistics, and is equipped with additional desirable properties that better preserve the data structure, which facilitate theoretical proofs and practical usages. In the experiments section, we provide a spectral clustering application and a testing independence application. All theorem proofs are in the appendix.

\section{Background} 
\label{sec:review}

\subsection*{Distance Correlation and Hilbert-Schmidt Independence Criterion}
Assume $(\mathbf{X},\mathbf{Y}) = \{(x_{i},y_{i}) \in \mathbb{R}^{p+q}, i=1,\ldots,n \}$ is the paired sample data where each pair $(x_{i},y_{i})$ is independently and identically distributed as $(X,Y) \sim F_{XY}$. Given a distance metric $d(\cdot,\cdot)$, let $\mathbf{D}^{\mathbf{X}}$ denote the $n \times n$ distance matrix of $\mathbf{X}$ such that $\mathbf{D}^{\mathbf{X}}_{ij}=d(x_i,x_j)$, $\mathbf{D}^{\mathbf{Y}}$ denote the corresponding distance matrix of $\mathbf{Y}$, $\mathbf{H}=\mathbf{I}-\frac{1}{n}\mathbf{J}$ denote the $n \times n$ centering matrix where $\mathbf{I}$ is the identity matrix and $\mathbf{J}$ is the matrix of ones. The biased sample distance covariance \cite{SzekelyRizzo2009} equals
\begin{align*}
&\Dcov_{n}^{b}(\mathbf{X},\mathbf{Y})= \frac{1}{n^2}trace(\mathbf{H}\mathbf{D}^{\mathbf{X}}\mathbf{H}\mathbf{H}\mathbf{D}^{\mathbf{Y}}\mathbf{H}),
\end{align*}
where the superscript $^{b}$ stands for a biased statistic. Note that it was called squared distance covariance in \cite{SzekelyRizzo2009}, and we drop the squared naming for ease of presentation in this paper.

Despite the elegant matrix formulation, the above statistic is biased, e.g., when $X$ is independent of $Y$, $E[\Dcov_{n}^{b}(\mathbf{X},\mathbf{Y})] > 0$ at any finite $n$. One strategy to cure the bias is to use a modified matrix $\mathbf{C}^{\mathbf{X}}$ \citep{SzekelyRizzo2014}:
\begin{align*}
\mathbf{C}^{\mathbf{X}}_{ij}=
 \begin{cases}
 \mathbf{D}^{\mathbf{X}}_{ij}-\frac{1}{n-2}\sum\limits_{t=1}^{n} \mathbf{D}^{\mathbf{X}}_{it}-\frac{1}{n-2}\sum\limits_{s=1}^{n} \mathbf{D}^{\mathbf{X}}_{sj}+\frac{1}{(n-1)(n-2)}\sum\limits_{s,t=1}^{n}\mathbf{D}^{\mathbf{X}}_{st}, \ i \neq j \\
 0, \mbox{ otherwise}.
 \end{cases}
\end{align*}
Similarly compute $\mathbf{C}^{\mathbf{Y}}$ from $\mathbf{D}^{\mathbf{Y}}$. Then the unbiased sample distance covariance equals:
\begin{align*}
& \Dcov_{n}(\mathbf{X}, \mathbf{Y}) = \frac{1}{n(n-3)}trace(\mathbf{C}^{\mathbf{X}}\mathbf{C}^{\mathbf{Y}})
\end{align*}
Distance correlation is the normalized distance covariance, computed as follows:
\begin{align}
\label{eq:dcov}
&\Dcor_{n}(\mathbf{X},\mathbf{Y})= \frac{\Dcov_{n}(\mathbf{X},\mathbf{Y})}{\sqrt{\Dcov_{n}(\mathbf{X},\mathbf{X})\Dcov_{n}(\mathbf{Y},\mathbf{Y})}} \in [-1,1].
\end{align}
Distance correlation is equivalent to distance covariance for testing purpose, but more advantageous for interpreting the test statistic, i.e., the correlation equals $1$ for linear relationship while the covariance is unbounded.

Suppose $(X^{'},Y^{'}), (X^{''},Y^{''})$ are two independent and identical copy of the random variable pair $(X,Y)$, the population distance covariance is defined as \cite{SzekelyRizzo2014}
\begin{align}
\label{eq1}
\Dcov(X,Y) & =E[d(X,X')d(Y,Y')]+E[d(X,X')]E[d(Y,Y')]-2E[d(X,X')d(Y,Y'')].
\end{align}
The sample statistic converges to the population version as sample size $n$ increases to infinity, and is also unbiased:
\begin{align*}
&\Dcov_{n}(\mathbf{X}, \mathbf{Y}) \stackrel{n \rightarrow \infty}{\rightarrow}  \Dcov(X,Y),\\
&E[\Dcov_{n}(\mathbf{X}, \mathbf{Y})]=\Dcov(X,Y).
\end{align*}
When $d(\cdot,\cdot)$ is the Euclidean metric, it was proved via characteristic functions \cite{SzekelyRizzo2009} that the population distance covariance equals $0$ if and only if $X$ and $Y$ are independent. Therefore, sample distance covariance converges to $0$ if and only if $X$ and $Y$ are independent, thus is universally consistent for testing independence. Note that $d(\cdot,\cdot)$ can be any metric not just the Euclidean, and the consistency property still holds for any strong negative type metric \cite{Lyons2013}. The convergence, unbiasedness, and consistent properties also hold for distance correlation by replacing $\Dcov$ by $\Dcor$. 

By replacing the distance metric $d(\cdot,\cdot)$ by a kernel function $k(\cdot,\cdot)$, and replacing the distance matrices $\mathbf{D}^{\mathbf{X}}$ and $\mathbf{D}^{\mathbf{Y}}$ by the respective kernel matrices $\mathbf{K}^{\mathbf{X}}$ and $\mathbf{K}^{\mathbf{Y}}$, the distance-based statistic become the respective kernel statistic. For example, when using the Gaussian kernel in the right hand side of Equation~\ref{eq1}, the left hand side becomes the population Hilbert-Schmidt covariance \cite{GrettonEtAl2005, SejdinovicEtAl2013}. Similarly one can define the biased sample Hilbert-Schmidt covariance, unbiased sample Hilbert-Schmidt covariance, unbiased correlation, and the population covariance. They are denoted by $\Hsic_{n}^{b}(\mathbf{X}, \mathbf{Y})$, $\Hsic_{n}(\mathbf{X}, \mathbf{Y})$, $\Hsicr_{n}(\mathbf{X}, \mathbf{Y})$, and $\Hsic(X,Y)$ respectively. When the kernel function $k(\cdot,\cdot)$ is characteristic \cite{GrettonEtAl2005}, $\Hsic_{n}(\mathbf{X}, \mathbf{Y})$ converges to $0$ if and only if $X$ and $Y$ are independent, thus is universally consistent. 

Hypothesis testing requires a test statistic and a p-value. Distance correlation is the test statistic, and a p-value is needed to decide whether one shall reject the independence hypothesis or not. Permutation test \cite{GoodPermutationBook} is the standard procedure for testing independence: compute $r$ permuted statistics by randomly permuting the samples of $\mathbf{Y}$ each time (i.e., when $\mathbf{Y}$ is a matrix of size $n \times p$, it corresponds to permute the rows), calculate the p-value as how often the sample statistic is larger than the permuted ones, and reject independence for small p-value. The permutation test is commonly used in almost every other method for testing independence \cite{HellerGorfine2013, Zhu2017,mgc2}, but can be slow since $r$ is at least $100$. Recently, fast chi-square test are made available for the unbiased distance or kernel correlation with validity and power guarantee \cite{fast1}, making the unbiased statistics more attractive for large data.

\subsection*{Metric and Kernel and Fixed-Point Transformation}
Here we introduce some required backgrounds of metric and kernel.
\begin{definition}
Let $\mathcal{Z}$ be a non-empty set. A metric $d(\cdot,\cdot): \mathcal{Z} \times \mathcal{Z} \rightarrow [0,\infty)$ is of negative type when it has the following property: for any $n \geq 2$, $x_{1},\ldots,x_{n} \in \mathcal{Z}$ and $a_{1},\ldots,a_{n} \in \mathbb{R}$ with $\sum_{i=1}^{n} a_{i}=0$, a negative type metric satisfies
\begin{align*}
\sum\limits_{i,j=1}^{n} a_{i}a_{j}d(x_{i},x_{j}) \leq 0.
\end{align*}
A strong negative type metric is a negative type metric with the following additional property: for any distribution $F_{X}$ and $F_{Y}$,
\begin{align}
\label{eq:metric1}
\int_{x'}\int_{x} d(x,x') \mathsf{d}(F_{X}(x)-F_{Y}(x))\mathsf{d}(F_{X}(x')-F_{Y}(x')) = 0 \mbox{ if and only if } F_{X}=F_{Y},
\end{align}
where $\mathsf{d}$ stands for the differential sign. 

A kernel $k(\cdot,\cdot):\mathcal{Z} \times \mathcal{Z} \rightarrow [0,\infty)$ is positive definite when for any $n \geq 2$, $x_{1},\ldots,x_{n} \in \mathcal{Z}$ and $a_{1},\ldots,a_{n} \in \mathbb{R}$, it holds that
\begin{align*}
\sum\limits_{i,j=1}^{n} a_{i}a_{j}k(x_{i},x_{j}) \geq 0.
\end{align*}
A characteristic kernel is a positive definite kernel with the following additional property: for any two random variables $X$ and $Y$ with distributions $F_{X}$ and $F_{Y}$,
\begin{align}
\label{eq:metric2}
E[k(\cdot,X)] = E[k(\cdot,Y)] \mbox{ if and only if } F_{X}=F_{Y}.
\end{align}
\end{definition}
Therefore, both the strong negative type metric and the characteristic kernel are able to uniquely characterize the distribution, which guarantees the consistency property of distance correlation and Hilbert-Schmidt independence criterion \cite{Lyons2013,GrettonEtAl2005}.

It was shown in \cite{SejdinovicEtAl2013} that for any given metric, there exists a fixed-point induced kernel such that distance covariance using the given metric equals Hilbert-Schmidt covariance using the induced kernel, and vice versa for any kernel and the induced metric. Note that this does not mean $\Hsic(X,Y)$ using Gaussian kernel equals $\Dcov(X,Y)$ using the Euclidean distance. It means there exists an induced metric of Gaussian kernel (or an induced kernel of Euclidean metric) such that these two population statistics are the same.
\begin{definition}
\label{defi1}
Let $\mathcal{Z}$ be a non-empty set.
For any metric $d(\cdot,\cdot): \mathcal{Z} \times \mathcal{Z} \rightarrow [0,\infty)$, 
its fixed-point induced kernel is defined as
\begin{align*}
\tilde{k}_{d}(x_{i},x_{j})=d(x_{i},z)+d(x_{j},z)-d(x_{i},x_{j}),
\end{align*}
at an arbitrary but fixed point $z$. For any kernel $k(\cdot,\cdot): \mathcal{Z} \times \mathcal{Z} \rightarrow [0,\infty)$, its induced metric is defined as 
\begin{align*}
\tilde{d}_{k}(x_{i},x_{j})=\frac{1}{2}k(x_{i},x_{i})+\frac{1}{2}k(x_{j},x_{j})-k(x_{i},x_{j}).
\end{align*}
\end{definition}
For ease of the equivalence presentation in this paper, the fixed-point transformation introduced here differs from the original definition in \cite{SejdinovicEtAl2013} by a factor of two. 

Next we introduce translation invariant metric and kernel \citep{Micchelli2006}. Many commonly used metrics and kernels are translation invariant, e.g., the Euclidean distance, $L^{p}$ norm, taxicab metric, any metric induced by a norm, the Gaussian kernel, the Laplacian kernel, etc.
\begin{definition}
Let $\mathcal{Z}$ be a non-empty set. A metric $d(\cdot,\cdot): \mathcal{Z} \times \mathcal{Z} \rightarrow [0,\infty)$ is translation invariant when there exists a function $g(\cdot)$ such that
\begin{align*}
d(x_{i},x_{j}) =g(x_i - x_j).
\end{align*}

Similarly, a kernel $k(\cdot,\cdot):\mathcal{Z} \times \mathcal{Z} \rightarrow [0,\infty)$ is translation invariant when there exists a function $g(\cdot)$ such that
\begin{align*}
k(x_{i},x_{j}) =g(x_i - x_j).
\end{align*}
\end{definition}

\section{Main Results}
\label{sec:main}
In this section, we introduce the bijection, prove a number of desirable properties, then proceed to show how the bijection establishes the equivalence between distance and kernel testing.

\subsection{The Bijective Transformation}
We propose the following bijection:
\begin{definition}
\label{def1}
Given sample data $\{x_i,i=1,\ldots,n\}$. For any metric $d(\cdot,\cdot)$, we define its bijective induced kernel as
\begin{align*}
\hat{k}_d(x_i,x_j)=\max\limits_{s,t \in [n]}(d(x_s,x_t))-d(x_i,x_j).
\end{align*}
For any kernel $k(\cdot,\cdot)$, we define the induced metric as
\begin{align*}
\hat{d}_k(x_i,x_j)=\max\limits_{s,t \in [n]}(k(x_s,x_t))-k(x_i,x_j).
\end{align*}
The subscripts $s,t \in [n]$ is a shorthand for $s=1,\ldots,n$ and $t=1,\ldots,n$.
\end{definition}
Note that the maximum element is always finite and well-defined for given sample data, but may not be fixed depending on the metric or kernel choice, making the bijection a data-adaptive transformation.

The proposed bijection is tightly related to the fixed-point transformation: the bijective induced kernel is a shift version of the fixed-point induced kernel that eliminates the fixed-point $z$; while the two induced metrics are essentially the same for most common kernels. 

\begin{restatable}{theorem}{one}
\label{thm1}
Given a metric $d(\cdot,\cdot)$ and a fixed point $z$, the bijective induced kernel and the fixed-point induced kernel are related via
\begin{align*}
\hat{k}_d(x_i,x_j)=\tilde{k}_d(x_i,x_j)+f(x_i)+f(x_j)
\end{align*}
for the shift function $f(x_{i})=\max\limits_{s,t \in [n]}(d(x_s,x_t))/2-d(x_{i},z)$.

Given a positive definite and translation invariant kernel $k(\cdot,\cdot)$, the bijective induced metric and the fixed-point induced metric are the same, i.e., for any $x_i, x_j$ we have
\begin{align*}
\hat{d}_k(x_i,x_j)=\tilde{d}_k(x_i,x_j).
\end{align*}
\end{restatable}

From now on, we always present the results for a given metric and the bijective induced kernel. All theorems established afterwards also hold for a given kernel and the bijective induced metric by replacing $\hat{k}_{d}(\cdot,\cdot)$ with $k(\cdot,\cdot)$ and $d(\cdot,\cdot)$ with $\hat{d}_{k}(\cdot,\cdot)$.

\begin{restatable}{theorem}{two}
\label{thm2}
The bijective induced kernel satisfies the following properties:

1. Non-negativity: $\hat{k}_{d}(x_i,x_j) \geq 0$ when $d(x_i,x_j) \geq 0$.

2. Identity: $\hat{k}_{d}(x_i,x_j)=\max\limits_{s,t \in [n]}\{\hat{k}_{d}(x_s,x_t)\}$ implies $x_i=x_j$, if and only if $d(x_i,x_j)=0$ implies $x_i=x_j$.

3. Symmetry: $\hat{k}_{d}(x_i,x_j)=\hat{k}_{d}(x_j,x_i)$ if and only if $d(x_i,x_j)=d(x_j,x_i)$.

4. Negative Type Metric to Positive Definite Kernel: the bijective induced kernel $\hat{k}_{d}(\cdot,\cdot)$ is positive definite if and only if $d(\cdot,\cdot)$ is of negative type.

5. Bijectivity: 
\begin{align*}
\hat{k}_{\hat{d}_{k}}(\cdot,\cdot)=k(\cdot,\cdot).
\end{align*}

6. Rank Preserving: 
\begin{align*}
d(x_i,x_s) < d(x_i,x_t) &\Rightarrow \hat{k}_{d}(x_i,x_s) >\hat{k}_{d}(x_i,x_t).
\end{align*}

7. Translation Invariant:
When $d(\cdot,\cdot)$ is translation invariant, $\hat{k}_{d}(\cdot,\cdot)$ is also translation invariant.
\end{restatable}

The first four properties are shared by both the bijection and fixed-point transformation. The last three properties are unique to the bijective induced kernel: bijectivity allows recovery of the original kernel and ensures no information is lost; rank preserving is important for any rank-based algorithm; and translation invariance is a necessity for follow-on inference to perform well such as in spectral clustering and support vector machine. 

Finally, note that we can alternatively define the bijection as:
\begin{align*}
&\hat{d}_k(x_i,x_j)=1-k(x_i,x_j)/\max\limits_{s,t \in [n]}(k(x_s,x_t)),\\
&\hat{k}_d(x_i,x_j)=1-d(x_i,x_j)/\max\limits_{s,t \in [n]}(d(x_s,x_t)),
\end{align*}
This is an equivalent definition up-to scaling by the maximum elements, and can be succinctly expressed in a matrix form:
\begin{align*}
&\hat{\mathbf{D}}_{\mathbf{K}}=\mathbf{J}-\mathbf{K}/\max(\mathbf{K}),\\
&\hat{\mathbf{K}}_{\mathbf{D}}=\mathbf{J}-\mathbf{D}/\max(\mathbf{D}).
\end{align*}
This alternative definition often facilitates implementation, computation, and visualization of the bijection, and all theoretical results in this manuscript still hold up-to a constant scaling. For ease of presentation we shall stick to Definition~\ref{def1} in this paper.

\subsection{The Equivalence for Hypothesis Testing}

The bijection enables distance covariance and Hilbert-Schmidt independence criterion to be always the same for testing, either between the biased methods or between the unbiased methods. Namely, when the metric and kernel being used are bijective of each other, the two methods have the same p-value in testing.
\begin{restatable}{theorem}{three}
\label{thm3}
Suppose distance covariance uses a given metric $d(\cdot,\cdot)$, and the Hilbert-Schmidt independence criterion $\hat{\Hsic}$ uses the bijective induced kernel $\hat{k}_{d}(\cdot,\cdot)$.

Given any sample data $(\mathbf{X},\mathbf{Y})$, it holds that
\begin{align*}
& \Dcov_{n}^{b}(\mathbf{X},\mathbf{Y})= \hat{\Hsic}_{n}^{b}(\mathbf{X},\mathbf{Y}),\\
& \Dcov_{n}(\mathbf{X},\mathbf{Y}) = \hat{\Hsic}_{n}(\mathbf{X},\mathbf{Y}) + O(\frac{1}{n^2}),
\end{align*}
where the remainder term $O(\frac{1}{n^2})$ is invariant to permutation.

Therefore, sample distance covariance and sample Hilbert-Schmidt covariance always yield the same p-value under permutation test.
\end{restatable}
The above theorem holds when the covariance terms are replaced by the correlations. It also implies that distance and kernel methods are equivalent for two-sample testing, see \cite{exact1}. There are two interesting observations from the proof: Firstly, if using the fixed-point transformation, the testing equivalence holds between the biased statistics but not between the unbiased statistics, i.e., the unbiased methods no longer share the same p-value when using the fixed-point transformation. Secondly, when using the bijective induced kernel, one can actually eliminate the invariant remainder term in the unbiased version by adding $\frac{1}{n-1}$ to all the off-diagonal entries of the modified matrices $\mathbf{C}^{\mathbf{X}}$ and $\mathbf{C}^{\mathbf{Y}}$.

\subsection{The Equivalence between Population Covariances}
Here we inspect the asymptotic properties of distance covariance and Hilbert-Schmidt independence criterion when using the bijection. The asymptotic properties need a separate treatment because the proposed bijection is data-adaptive in nature: the maximum element may increase to infinity as sample size increases. To that end, we define the population Hilbert-Schmidt covariance using a bijective induced kernel as the limit of sample covariance, and establish the equivalence between the population version as follows:
\begin{restatable}{theorem}{four}
\label{thm4}
Given a random variable pair $(X,Y)$ and a given metric $d(\cdot,\cdot)$ where the population distance covariance is well-defined as in Equation~\ref{eq1}. For the bijective induced kernel $\hat{k}_{d}(\cdot,\cdot)$, we define the population Hilbert-Schmidt covariance as the limit of sample covariance:
\begin{align*}
\hat{\Hsic}(X,Y)=\lim_{n \rightarrow \infty }\hat{\Hsic}_{n}(\mathbf{X},\mathbf{Y}),
\end{align*}
Then the population Hilbert-Schmidt covariance equals the population distance covariance, i.e., 
\begin{align*}
&\hat{\Hsic}(X,Y)=\Dcov(X,Y)\\
&=E[d(X,X')d(Y,Y')]+E[d(X,X')]E[d(Y,Y')]-2E[d(X,X')d(Y,Y'')].
\end{align*}
\end{restatable}
This leads to the next theorem:

\begin{restatable}{theorem}{five}
\label{thm5}
When the given metric $d(\cdot,\cdot)$ is of strong negative type, the bijective induced kernel $\hat{k}_{d}(\cdot,\cdot)$ is asymptotically a characteristic kernel. When the given kernel $k(\cdot,\cdot)$ is characteristic, the bijective induced metric $\hat{d}_{k}(\cdot,\cdot)$ is asymptotically of strong negative type.
\end{restatable}

\section{Experiments}
\label{sec:simu}
\subsection{The Equivalence Simulations}
\begin{figure}
\includegraphics[width=0.9\textwidth,trim={0.5cm 0cm 0cm 0cm},clip]{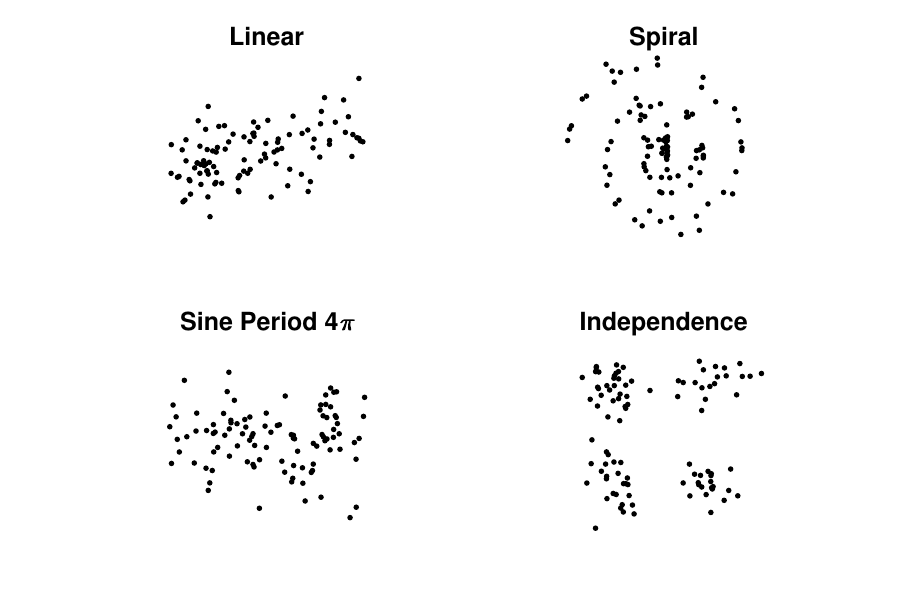}
\caption{Visualize linear, spiral, sine, and independent cloud relationships at $n=100$ with noise.}
\label{fig4}
\end{figure}

Here we illustrate the testing equivalence via four representative simulations (linear, spiral, sine, and independent cloud) at $n=100$ and $p=q=1$ with some white noise. The exact functions can be found in appendix of \cite{mgc2}, and a visualization is provided in Figure~\ref{fig4}. For each relationship, we generate $(\mathbf{X},\mathbf{Y})$, compute the distance correlation using Euclidean distance, compute the Hilbert-Schmidt correlation using the bijective induced kernel, and calculate their p-values under permutation test with $r=100$ random permutations (same permutation applied to every method). The results are reported in Table~\ref{table1} for both the biased and unbiased methods. 

As expected, the biased statistics are always the same, while the unbiased statistics are almost the same. The p-values are strictly the same between the biased methods, or between the unbiased methods. This validates that the bijection preserves the testing equivalence on sample data. Note that the actual numbers will be different each time due to random data generation, or as the sample size, dimensionality, dependency type changes. Nevertheless, the testing equivalence is always strict and holds between distance correlation and Hilbert-Schimidt independence criterion when using the bijection.

\begin{table*}[htbp]
\centering
\caption{Sample Statistic and P-value using Distance Correlation and Hilbert-Schmidt Independence Criterion with Bijective Induced Metric. The last four rows are unbiased methods while the previous four rows are biased methods.}
\label{table1}
\begin{tabular}{|c||c|c|c|c|}
\hline
$(\mathbf{X},\mathbf{Y})$ & Linear & Spiral & Sine & Independence \\
\hline
$\Dcor_{n}^{b}$ & $0.1095$  & $0.0351$ & $0.0323$ & $0.0091$ \\
\hline
$p(\Dcor_{n}^{b})$ & $0.01$  & $0.31$ & $0.22$ & $0.80$ \\
\hline
$\hat{\Hsicr}_{n}^{b}$ & $0.1095$  & $0.0351$ & $0.0323$ & $0.0091$ \\
\hline
$p(\hat{\Hsicr}_{n}^{b})$ & $0.01$  & $0.31$ & $0.22$ & $0.80$ \\
\hline
$\Dcor_{n}$ & $0.0860$  & $0.0200$ & $0.0066$ & $-0.0092$ \\
\hline
$p(\Dcor_{n})$ & $0.01$  & $0.24$ & $0.25$ & $0.78$ \\
\hline
$\hat{\Hsicr}_{n}$ & $0.0892$  & $0.0138$ & $0.0066$ & $-0.0078$ \\
\hline
$p(\hat{\Hsicr}_{n})$ & $0.01$  & $0.24$ & $0.25$ & $0.78$ \\
\hline
\end{tabular}
\end{table*}

\subsection{Spectral Clustering Application}
\begin{figure}
\centering
\includegraphics[width=1.0\textwidth,trim={0.7cm 0cm 0cm 0cm},clip]{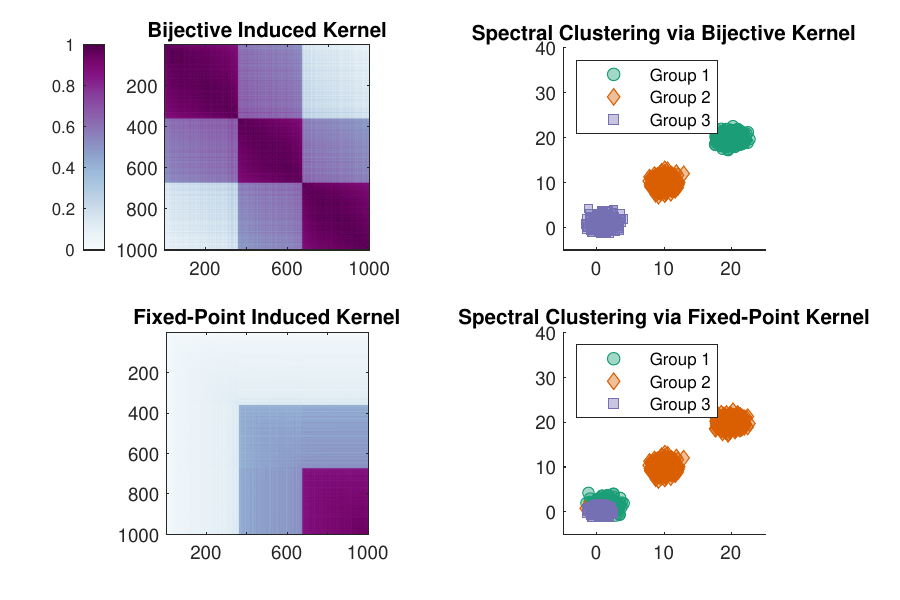}
\caption{Generate $\{w_i\}$ from a 2D Gaussian mixture of three components at $n=1000$. Starting with Euclidean distance, the first row visualizes the bijective induced kernel matrix and the spectral clustering result, and the second row visualizes the fixed-point induced kernel matrix and the spectral clustering result. }
\label{fig1}
\end{figure}

In this section, we use spectral clustering to demonstrate the advantage of the bijection. Spectral clustering uses the eigen-decomposition of a kernel matrix to perform dimensionality reduction then clustering. It is popular in image segmentation \citep{ShiMalik2000,Jordan2002,Luxburg2007} and a suitable benchmark to compare the quality of a kernel transformation. We set $n=1000$ and generate $w_i$ from a two-dimensional Gaussian mixture with three equally likely components. Each component has a different mean with identity matrix as the covariance. We compute the Euclidean distance matrix of $\{w_i\}$, transform it to the bijective induced kernel and the fixed-point induced kernel, and apply the spectral clustering algorithm from  \cite{Jordan2002} to both kernel matrices. The top row of Figure~\ref{fig1} shows that the bijective induced kernel exhibits a clear block structure and is able to produce a perfect clustering result upon applying spectral clustering. The bottom row shows a somewhat distorted kernel matrix caused by the fixed-point (and apparently not rank-preserving nor translation-invariant), which causes incorrect groupings for many observations when using spectral clustering. 

\subsection{Multiscale Graph Correlation for Kernel}
\begin{figure}
\includegraphics[width=1.0\textwidth,trim={1cm 0cm 0cm 0cm},clip]{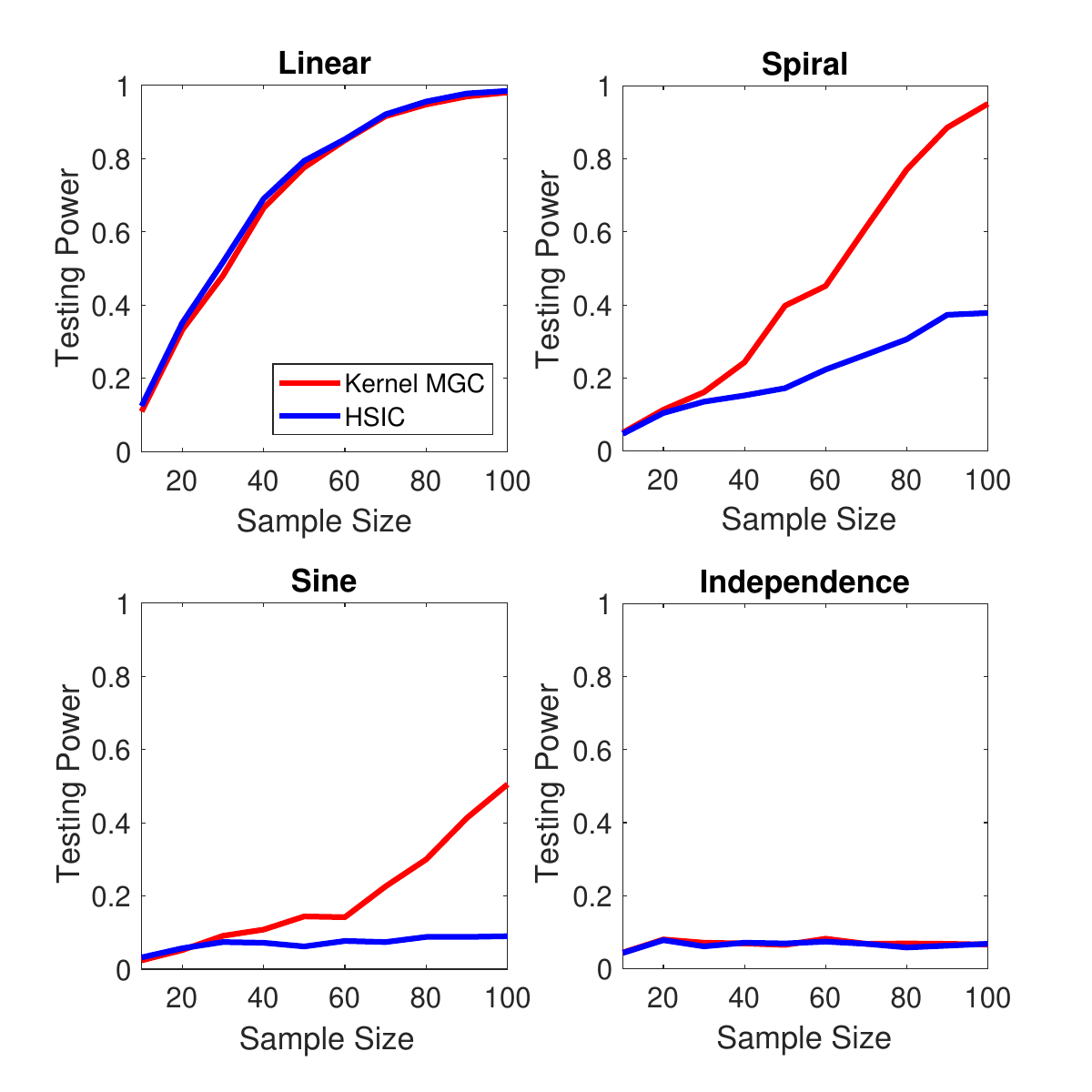}
\caption{Comparing testing power of Hilbert-Schmidt independence criterion and Kernel multiscale graph correlation for four simulations.}
\label{fig5}
\end{figure}

 As another application, we show how the multiscale graph correlation can be directly applied to a kernel via the bijection. The original multiscale graph correlation is the optimal local distance correlation via nearest-neighbor, which is applicable to any distance metric and exhibits a better testing power over distance correlation \cite{mgc1,mgc2}. The bijection enables a straightforward kernel version: given any kernel, apply the multiscale graph correlation to the bijective induced metric, then test independence by permutation test. We compare the Hilbert-Schmidt independence criterion (which uses the Gaussian kernel with median distance as the bandwidth) with the multiscale graph correlation (using the bijective induced metric of the same Gaussian kernel), and evaluate their testing power for the four simulations in Figure~\ref{fig4} at $n=100$. For each simulation, we repeatedly generate sample data for $1000$ times, count how often each method correctly identifies the relationship, and compute the testing power of each statistic at $\alpha=0.05$ using $1000$ permutations. Figure~\ref{fig5} shows the kernel multiscale graph correlation is able to improve the testing power for strongly nonlinear dependencies, exhibits almost the same power as Hilbert-Schmidt independence criterion for linear dependency, and correctly controls the type 1 error level for the independence relationship. Note that multiscale graph correlation cannot be directly applied to a fixed-point induced metric, as the nearest-neighbor structure of each observation is not preserved during the fixed-point transformation as evidenced from the left two panels of Figure~\ref{fig1}.

\section*{Acknowledgment}
This work was graciously supported by the Microsoft Research,
%
the National Science Foundation award DMS-1921310,
%
%
%
the Defense Advanced Research Projects Agency's (DARPA) SIMPLEX program through SPAWAR contract N66001-15-C-4041, and DARPA Lifelong Learning Machines program through contract FA8650-18-2-7834.
%
%
%
The authors thank Dr. Minh Tang, Dr. Carey Priebe, and Dr. Franca Guilherme for their comments and suggestions. We thank Dr. Arthur Gretton and anonymous feedback regarding the asymptotic property of the bijection. We further thank the journal editor and reviewers for their valuable suggestions that significantly improved the organization and exposition of the paper.

\bibliographystyle{chicago}
\bibliography{mgc}

\clearpage
\appendix
\setcounter{figure}{0}
\renewcommand{\thefigure}{E\arabic{figure}}
\renewcommand{\thesubsection}{\thesection.\arabic{subsection}}
\renewcommand{\thesubsubsection}{\thesubsection.\arabic{subsubsection}}
\pagenumbering{arabic}
\renewcommand{\thepage}{\arabic{page}}

\bigskip
\begin{center}
{\large\bf APPENDIX}
\end{center}



\section{Proofs}
\label{sec:proofs}

\one*
\begin{proof}
The first part from metric to kernel follows easily by algebraic manipulation. To prove the second part where the two induced metrics are the same, it suffices to show that for any sample data $\{x_i\}$ and any $i \in [n]$, it holds that
\begin{align*}
k(x_{i},x_{i})=\max\limits_{s,t \in [n]}(k(x_s,x_t))
\end{align*}
Note that because of translation invariance, $k(x_{i},x_{i})=k(x_{j},x_{j})$ for any $i \neq j$, so $k(x_{i},x_{i})$ must be the same maximum element for any $i$.

We prove by contradiction: suppose $k(x_{i},x_{i})$ is not the maximum element. Thus there must exist some $i$ and $j \neq i$ such that $k(x_{i},x_{j}) > k(x_{i},x_{i})$. Denote $\mathbf{K}$ as the $n \times n$ kernel matrix of sample data $\{x_i\}$. Let $a$ be the $1 \times n$ vector with $1$ at $i$th entry, $-1$ at $j$th entry, and zero elsewhere. Then 
\begin{align*}
a \mathbf{K} a^{T}=k(x_{i},x_{i})+k(x_{j},x_{j})-2k(x_{i},x_{j}) < 0.
\end{align*}
This contradicts the kernel being positive definite. Therefore, $k(x_{i},x_{i})$ must be the maximum element, and the two induced metrics are the same.
\end{proof}

\two*
\begin{proof}
As all other properties are straightforward to verify, here we only prove property (4). Without loss of generality, we assume the maximum metric or kernel element is always $1$, since the property is invariant to constant scaling. Denote $\mathbf{a}=[a_{1},\ldots,a_{n}]$ as a row vector, its mean as $\frac{1}{n}\sum_{i=1}^{n} a_{i}= \bar{a}$, $\mathbf{1}$ as the column vector of ones, and 
$\mathbf{b}$ as a zero-mean vector. Note that $\mathbf{a}-\bar{a} \mathbf{1}^{T}$ is always a zero-mean vector. Given sample data $\{x_i, i=1,\ldots,n\}$, denote $\mathbf{D}$ as the $n \times n$ distance matrix, $\hat{\mathbf{K}}_{\mathbf{D}}$ as the $n \times n$ bijective induced kernel matrix, and recall $\mathbf{J}$ is the matrix of ones.

To prove the only if direction, it suffices to show that when $\mathbf{a} \hat{\mathbf{K}}_{\mathbf{D}}\mathbf{a}^{T} \geq 0$ for any vector $\mathbf{a}$, it holds that $\mathbf{b}\mathbf{D}\mathbf{b}^{T} \leq 0$ for any zero-mean vector $\mathbf{b}$. It follows that
\begin{align*}
\mathbf{b}\mathbf{D}\mathbf{b}^{T}=\mathbf{b}(\mathbf{J}-\hat{\mathbf{K}}_{\mathbf{D}})\mathbf{b}^{T} =\mathbf{b}(-\hat{\mathbf{K}}_{\mathbf{D}})\mathbf{b}^{T} \leq 0.
\end{align*}
where $\mathbf{D}=\mathbf{J}-\hat{\mathbf{K}}_{\mathbf{D}}$ is the bijection, $\mathbf{b}\mathbf{J}\mathbf{b}^{T}=0$, and the last inequality follows from observing that $-\mathbf{a} \hat{\mathbf{K}}_{\mathbf{D}}\mathbf{a}^{T} \leq 0$ and $\mathbf{b}$ is a special case of $\mathbf{a}$.

To prove the if direction, it suffices to show that when $\mathbf{b}\mathbf{D}\mathbf{b}^{T} \leq 0$ for any zero-mean vector $\mathbf{b}$, it holds that $\mathbf{a}(\mathbf{J}-\mathbf{D})\mathbf{a}^{T} \geq 0$ for any vector $\mathbf{a}$. This is established by the following:
\begin{align*}
& (\mathbf{b}+ \bar{a}\mathbf{1}^{T})(\mathbf{J}-\mathbf{D})(\mathbf{b}+ \bar{a}\mathbf{1}^{T})^{T} \\
= & \ \bar{a}^2\mathbf{1}^{T} \mathbf{J}\mathbf{1}-\mathbf{b} \mathbf{D}\mathbf{b}^{T}- 2\bar{a}\mathbf{1}^{T} \mathbf{D}\mathbf{b}^{T} - \bar{a}^2\mathbf{1}^{T} \mathbf{D}\mathbf{1}\\
= & \ -\mathbf{b} \mathbf{D}\mathbf{b}^{T}+\bar{a}^2\mathbf{1}^{T} \mathbf{J}\mathbf{1}- 2\bar{a}\mathbf{1}^{T} \mathbf{D}\mathbf{a}^{T} + \bar{a}^2\mathbf{1}^{T} \mathbf{D}\mathbf{1} \\
= & \ -\mathbf{b} \mathbf{D}\mathbf{b}^{T}+n^2 \bar{a}^2 - 2\bar{a}\sum\limits_{i=1}^{n} \{a_i \sum\limits_{j=1}^{n}\hat{d}_{k}(x_i,x_j)\} + \bar{a}^2\sum\limits_{i,j=1}^{n} \hat{d}_{k}(x_i,x_j)\\
\geq & \ -\mathbf{b} \mathbf{D}\mathbf{b}^{T} \geq 0,
\end{align*}
where the first equality follows by expanding all terms and eliminating any term containing $\mathbf{b}\mathbf{1}=0$; the second equality follows by noting that $\mathbf{b}=(\mathbf{a}- \bar{a}\mathbf{1}^{T})$ and expand the term $2\bar{a}\mathbf{1}^{T} \mathbf{D}\mathbf{b}^{T}$ accordingly; the third equality decomposes the matrix notations into summations for every term except $\mathbf{b} \mathbf{D}\mathbf{b}^{T}$. The last line follows because $-\mathbf{b} \mathbf{D}\mathbf{b}^{T} \geq 0$, and the other three terms on the fourth line is no smaller than $0$:
\begin{align*}
& n^2 \bar{a}^{2} - 2\bar{a}\sum\limits_{i=1}^{n} \{a_i \sum\limits_{j=1}^{n}d(x_i,x_j)\} + \bar{a}^2\sum\limits_{i,j=1}^{n} d(x_i,x_j) \\
= & \ n^2 \bar{a}^{2} - \bar{a} \sum\limits_{i=1}^{n} \{(2a_i -\bar{a}) \sum\limits_{j=1}^{n}d(x_i,x_j)\}\\
\geq & \ n^2 \bar{a}^{2} - n\bar{a} \sum\limits_{i=1}^{n} (2a_i-\bar{a}) \\
= & \ 2n^2 \bar{a}^2 - 2n^2 \bar{a}^2 =0
\end{align*}
The third lines follows by noting that $d(x_i,x_j) \leq 1$ because the maximum kernel element is assumed to be $1$. 
\end{proof}

\three*
\begin{proof}
It suffices to prove the equivalence of the two covariance terms. Then the permuted statistics using distance covariance and Hilbert-Schmidt covariance are always the same (using the same permutation), leading to same p-value using permutation test.

Without loss of generality, assume $\max\limits_{s,t \in [1,n]}(d(x_s,x_t))=\max\limits_{s,t \in [1,n]}(d(y_s,y_t))=1$. Given sample data $\mathbf{X}$, we denote $\mathbf{D}^{\mathbf{X}}$ as the distance matrix, the bijective induced kernel matrix as $\hat{\mathbf{K}}_{\mathbf{D}}^{\mathbf{X}}$, and similarly for $\mathbf{Y}$. The equivalence for the biased covariances follows as:
\begin{align*}
& \Dcov_{n}^{b}(\mathbf{X},\mathbf{Y}) - \hat{\Hsic}_{n}^{b}(\mathbf{X},\mathbf{Y}) \\ 
= & \ \frac{1}{n^2}\{ trace(\mathbf{D}^{\mathbf{X}}\mathbf{H}\mathbf{D}^{\mathbf{Y}}\mathbf{H})- trace(\hat{\mathbf{K}}_{\mathbf{D}}^{\mathbf{X}}\mathbf{H}\hat{\mathbf{K}}_{\mathbf{D}}^{\mathbf{Y}}\mathbf{H}) \}\\ 
=& \ \frac{1}{n^2}\{ trace(\mathbf{D}^{\mathbf{X}}\mathbf{H}\mathbf{D}^{\mathbf{Y}}\mathbf{H})- trace((\mathbf{J}-\mathbf{D}^{\mathbf{X}})\mathbf{H}(\mathbf{J}-\mathbf{D}^{\mathbf{Y}})\mathbf{H}) \}\\
=& \ \frac{1}{n^2}\{ trace(\mathbf{D}^{\mathbf{X}}\mathbf{H}\mathbf{J}\mathbf{H}+\mathbf{D}^{\mathbf{Y}}\mathbf{H}\mathbf{J}\mathbf{H}-\mathbf{J}\mathbf{H}\mathbf{J}\mathbf{H}) \}\\
= & \ 0,
\end{align*}
where the last equality ie because of
\begin{align*}
\mathbf{J}\mathbf{H}=\mathbf{J}(\mathbf{I}-\frac{1}{n}\mathbf{J})=\mathbf{J}-\mathbf{J}=0.
\end{align*}

To prove the unbiased version, we denote the modified matrices as $\mathbf{C}^{\mathbf{X}}$ and $\hat{\mathbf{C}}^{\mathbf{X}}$ respectively corresponding to the distance matrix $\mathbf{D}^{\mathbf{X}}$ and the induced kernel matrix $\hat{\mathbf{K}}_{\mathbf{D}}^{\mathbf{X}}$. As their diagonal entries are always $0$, it suffices to analyze the off-diagonal entries of $\mathbf{C}^{\mathbf{X}}$ and $\hat{\mathbf{C}}^{\mathbf{X}}$: for each $i \neq j$,
\begin{align*}
\hat{\mathbf{C}}^{\mathbf{X}}(i,j)&=
 \hat{\mathbf{K}}^{\mathbf{X}}_{\mathbf{D}}(i,j)-\frac{1}{n-2}\sum\limits_{t=1}^{n} \hat{\mathbf{K}}^{\mathbf{X}}_{\mathbf{D}}(i,t)\\
 &-\frac{1}{n-2}\sum\limits_{s=1}^{n} \hat{\mathbf{K}}^{\mathbf{X}}_{\mathbf{D}}(s,j)+\frac{1}{(n-1)(n-2)}\sum\limits_{s,t=1}^{n}\hat{\mathbf{K}}^{\mathbf{X}}_{\mathbf{D}}(s,t) \\
 & = (1-\mathbf{D}^{\mathbf{X}}(i,j))-\frac{1}{n-2}\sum\limits_{t=1}^{n} (1-\mathbf{D}^{\mathbf{X}}(i,t))\\
 &-\frac{1}{n-2}\sum\limits_{s=1}^{n} (1-\mathbf{D}^{\mathbf{X}}(s,j))+\frac{1}{(n-1)(n-2)}\sum\limits_{s,t=1}^{n}(1-\mathbf{D}^{\mathbf{X}}(s,t)) \\
  & = -(\mathbf{C}^{\mathbf{X}}(i,j)+\frac{1}{n-1}).
\end{align*}
Therefore, the unbiased sample Hilbert-Schmidt covariance satisfies
\begin{align*}
\hat{\Hsic}_{n}(\mathbf{X}, \mathbf{Y}) &= \frac{1}{n(n-3)}trace(\hat{\mathbf{C}}^{\mathbf{X}}\hat{\mathbf{C}}^{\mathbf{Y}}) \\
&= \frac{1}{n(n-3)}trace((\mathbf{C}^{\mathbf{X}}+\frac{\mathbf{J}-\mathbf{I}}{n-1})(\mathbf{C}^{\mathbf{Y}}+\frac{\mathbf{J}-\mathbf{I}}{n-1})) \\
&= \Dcov_{n}(\mathbf{X}, \mathbf{Y}) + \frac{trace((n-1)\mathbf{C}^{\mathbf{X}}(\mathbf{J}-\mathbf{I})+(n-1)\mathbf{C}^{\mathbf{Y}}(\mathbf{J}-\mathbf{I})+(\mathbf{J}-\mathbf{I})^2)}{n(n-1)^2(n-3)}\\
&= \Dcov_{n}(\mathbf{X}, \mathbf{Y}) + O(\frac{1}{n^2})
\end{align*}
As the remainder term is invariant to permutation, the p-value is always the same between $\hat{\Hsic}_{n}(\mathbf{X}, \mathbf{Y})$ and $\Dcov_{n}(\mathbf{X}, \mathbf{Y})$. As evident by the second line above, the two unbiased sample covariances can be made exactly the same by adding $\frac{1}{n}$ to the modified matrices $\hat{\mathbf{C}}^{\mathbf{X}}$ and $\hat{\mathbf{C}}^{\mathbf{Y}}$.

Note that one can replicate the equivalence proof between the biased covariances using the fixed-point induced kernel. However, such equivalence does not hold between the unbiased covariances using the fixed-point induced kernel: as the fixed-point terms no longer cancel out, the remainder term is dependent on the choice of fixed-point and no longer invariant to permutation.
\end{proof}

\four*
\begin{proof}
Denote
\begin{align*}
\max\limits_{s,t \in [1,n]}(d(x_s,x_t))&=a_n \\ \max\limits_{s,t \in [1,n]}(d(y_s,y_t))&=b_n.
\end{align*}
Based on the definition of $\hat{\Hsic}(X,Y)$ for induced kernel and the sample equivalence so far, it follows that
\begin{align*}
&\hat{\Hsic}(X,Y) - \Dcov(X,Y)\\
=& \ \lim_{n \rightarrow \infty }\{\hat{\Hsic}_{n}(\mathbf{X},\mathbf{Y}) -\Dcov_{n}(\mathbf{X},\mathbf{Y})\} \\
=& \ \lim_{n \rightarrow \infty } \frac{1}{n^2} \{trace(\hat{\mathbf{K}}^{\mathbf{X}}_{\mathbf{D}}\mathbf{H}\hat{\mathbf{K}}^{\mathbf{Y}}_{\mathbf{D}}\mathbf{H})-trace(\mathbf{D}^{\mathbf{X}}\mathbf{H} \mathbf{D}^{\mathbf{Y}}\mathbf{H})\} \\
=& \ \lim_{n \rightarrow \infty } \frac{1}{n^2} \{trace((a_n \mathbf{J}- \mathbf{D}^{\mathbf{X}})\mathbf{H}(b_n \mathbf{J}- \mathbf{D}^{\mathbf{Y}})\mathbf{H} - \mathbf{D}^{\mathbf{X}}\mathbf{H} \mathbf{D}^{\mathbf{Y}}\mathbf{H})\} \\
=& \ \lim_{n \rightarrow \infty } \frac{1}{n^2} \{trace(a_n b_n \mathbf{J}\mathbf{H}\mathbf{J}\mathbf{H}- b_n\mathbf{D}^{\mathbf{X}}\mathbf{H}\mathbf{J}-a_n\mathbf{J}\mathbf{H}\mathbf{D}^{\mathbf{Y}} \}\\
=& \ 0.
\end{align*}
Namely, the maximum elements always cancel out, which does not affect the limiting equivalence even if the maximum increases to infinity. Therefore, it holds that
\begin{align*}
&\hat{\Hsic}(X,Y) = \Dcov(X,Y)\\
&= E[k(X,X')k(Y,Y')]+E[k(X,X')]E[k(Y,Y')]-2E[k(X,X')k(Y,Y'')],
\end{align*}
and the population Hilbert-Schmidt covariance using the induced kernel is always well-defined and equals the population distance covariance.
\end{proof}

\five*
\begin{proof}
Given any characteristic kernel $k(\cdot,\cdot)$, the sample \Hsic~converges to $0$ if and only if $X$ and $Y$ are independent. Then distance covariance using the bijective induced metric $\hat{d}_{k}(\cdot,\cdot)$ is exactly the same and also converges to $0$ if and only if $X$ and $Y$ are independent by Theorem~\ref{thm4}. Thus the induced metric must be asymptotically of strong negative type by \cite{Lyons2013}. Conversely, when $d(\cdot,\cdot)$ is of strong negative type, $\hat{k}_{d}(\cdot,\cdot)$ must be asymptotically characteristic.
\end{proof}

\end{document}